# Decision Support Systems Using Intelligent Paradigms

**Cong Tran and Lakhmi Jain**
School of Electrical and Information Engineering, University of South Australia,
Email: Lakhmi.Jain@unisa.edu.au

**Ajith Abraham**
Department of Computer Science, Oklahoma State University, 700 N Greenwood Avenue,
Tulsa, OK 74106, USA, Email: ajith.abraham@ieee.org

**Abstract:** Decision-making is a process of choosing among alternative courses of action for solving complicated problems where multi-criteria objectives are involved. The past few years have witnessed a growing recognition of Soft Computing (SC) technologies that underlie the conception, design and utilization of intelligent systems. In this paper, we present different SC paradigms involving an artificial neural network trained using the scaled conjugate gradient algorithm, two different fuzzy inference methods optimised using neural network learning/evolutionary algorithms and regression trees for developing intelligent decision support systems. We demonstrate the efficiency of the different algorithms by developing a decision support system for a Tactical Air Combat Environment (TACE). Some empirical comparisons between the different algorithms are also provided.

## 1. Introduction

Several decision support systems have been developed mostly in various fields including medical diagnosis, business management, control system, command and control of defence and air traffic control and so on [7][21]. Usually previous experience or expert knowledge is often used to design decision support systems. The task becomes interesting when no prior knowledge is available. The need for an intelligent mechanism for decision support comes from the well-known limits of human knowledge processing. It has been noticed that the need for support for human decision makers is due to four kinds of limits: cognitive, economic, time and competitive demands [7]. Several adaptive learning frameworks for constructing intelligent decision support systems have been proposed [4][20]]. Figure 1 summarizes the basic functional aspects of a decision support system. A database is created from the available data and human knowledge. The learning process then builds up the decision rules. The developed rules are further fine tuned depending upon the quality of the solution using a supervised learning process.

To develop an intelligent decision support system, we need a holistic view on the various tasks to be carried out including data management and knowledge management (reasoning techniques). The focus of this paper is knowledge management, which consists of facts and inference rules used for reasoning.

Fuzzy logic, when applied to decision support systems, provides formal methodology to capture valid patterns of reasoning about uncertainty. Neural networks are popularly known as blackbox function approximators. The recent research work showing the capabilities of rule extraction [17] from a trained network positions neurocomputing as a good decision support tool. Recently Evolutionary Computation (EC) has been successful as a powerful global optimisation tool due to the success in several problem domains. EC works by

Simulating evolution on a computer by iterative generation and alteration processes operating on a set of candidate solution that forms a population. Due to the complementarity of neural networks, fuzzy inference systems and evolutionary computation, the recent trend is to fuse various systems to form a more powerful integrated system, to overcome their individual weakness. Decision trees [3] have emerged as a powerful machine learning technique due to a simple, apparent, and fast reasoning process. Decision trees can be related to artificial neural networks by mapping them into a class of artificial neural networks or entropy nets with far fewer connections.

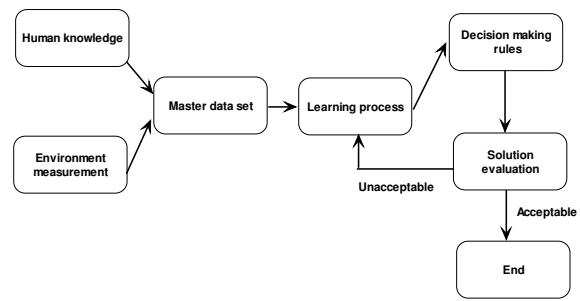

**Figure 1. Intelligent decision support system**

In Section 2, we present the complexity of the Tactical Air Combat Decision Support Systems (TACDSS) followed by some theoretical foundation on neural networks, fuzzy inference systems, neuro-fuzzy systems and decision trees in Section 3. In Sections 4 and 5, we present the different adaptation procedures for optimising fuzzy inference systems. A Takagi-Sugeno and a Mamdani fuzzy inference system learned using neural network learning techniques and evolutionary computation is discussed. Experimentation results using the different connectionist paradigms are presented in Section 6. A detailed discussion about the different

experimental results are given in Section 7 followed by conclusion towards the end

## 2. Tactical Air Combat Decision Support Systems

Implementation of a reliable decision support system involves two important factors: collection and analysis of prior information and the evaluation of the solution. The data could be an image or a pattern, real number, binary code or natural language text data depending on the objects of the problem environment. An object of the decision problem is also known as the decision factor. These objects can be expressed mathematically in the decision problem domain as a universal set where the decision factor is a set and decision data is an element of a set. The decision factor is a subset of the decision problem. If we can call the Decision Problem (DP) as *X* and the decision factor (DF) as '*A*' then the decision data (DD) could be labelled as '*a*'. Suppose the set *A* has members $a_1, a_2, ... , a_n$ then the set *A* can be denoted by $A = \{a_1, a_2, ..., a_n\}$ or can be written as:

$$A = \{a_i | i \in R_n\} \quad (1)$$

where *i* is called *set index*, the symbol '|' is read as 'such that' and $R_n$ is the set of *n* real numbers. A subset '*A*' of *X*, denoted $A \subseteq X$, is a set of elements that is contained within the universal set *X*. For optimal decision-making, the system should be able to adaptively process the information provided by words or any natural language description of the problem environment.

To illustrate the proposed approach, we considered a case study based on a tactical environment problem. We aim to develop a tactical environment decision support system for a pilot or mission commander in tactical air combat. We will attempt to present the complexity of the problem with some scenarios of the problem. In Figure 2 a typical scenario of air combat tactical environment is presented. The Airborne Early Warning and Control (AEW&C) is performing surveillance in a particular area of operation. It has two hornets (F/A-18s) under its control at the ground base as shown "+" in the left corner of Figure 2. An air-to-air fuel tanker (KB707) "□" is on station and the location and status are known to the AEW&C. One of the hornets is on patrol in the area of Combat Air Patrol (CAP). Sometime later, the AEW&C on-board sensors detect a hostile aircraft that is shown in "○". When the hostile aircrafts enter the surveillance region (shown as dashed circle) the mission system software is able to identify the enemy aircraft and its distance from the Hornets in the ground base or in the CAP.

The mission operator has few options to make a decision on the allocation of hornets to intercept the enemy aircraft.

- Send the Hornet directly to the spotted area and intercept
- Call the Hornet in the area back to ground base and send another Hornet from the ground base
- Call the Hornet in the area for refuel before intercepting the enemy aircraft

The mission operator will base his/her decisions on a number of decision factors, such as:
- Fuel used and weapon status of hornet in the area
- Interrupt time of Hornet in the ground base in the Hornet at the CAP to stop the hostile.
- The speed of the enemy fighter aircraft and the type of weapons it possesses.

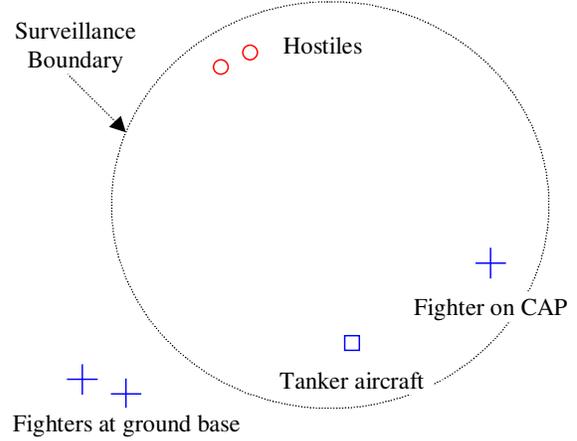

Figure 2. A typical scenario of air combat

Table 1. Decision factors for the tactical air combat

| Fuel used | Time Intercept | Weapon Status | Danger Situation | Evaluation Plan |
|---|---|---|---|---|
| Full | Fast | Sufficient | Very Danger | Good |
| Half | Normal | Enough | Danger | Acceptable |
| Low | Slow | Insufficient | Endanger | Bad |

From the above scenario, it is evident that there are important decision factors of the tactical environment that might directly affect the air combat decision. For demonstrating our proposed approach, we will simplify the problem by handling only a few important decision factors such as "fuel status", "weapon possession status" and "interrupt time" (Hornet in the ground base and the Hornet in the area of CAP) and the "Situation Awareness". These factors are tabulated in Table 1. The developed tactical air combat decision rules should be able to incorporate all the above-mentioned decision factors.

### Knowledge of Tactical Air Combat Environment

How human knowledge could be extracted to a database? Very often people express knowledge as natural language (spoken language) or using letters or symbolic terms. The human knowledge can be analysed and converted into an information table. There are several methods to extract human knowledge. Some researchers use the Cognitive Work Analysis (CWA) [16] and the Cognitive Task Analysis (CTA) [13]. The CWA is a technique to analyse, design and evaluate the human computer interactive systems. The CTA is a method to identify cognitive skill, mental demands and needs to perform task proficiency. The CTA focuses on describing the representation of the cognitive elements that defines goal generation and decision making. The CTA is a reliable method to extract the human knowledge because it is based on the observations or an interview. We have used the CTA technique to set up the expert knowledge base to formulate a key knowledge for building the complete decision support system. For the Tactical Air Combat Environment (TACE) discussed before, we have four decision factors that could affect the final decision options of "hornet in the CAP" or "hornet at the ground base". These are "fuel status" that is the quantity of fuel available to perform the intercept, the "weapon possession status" presenting the state of available weapons inside the hornet, the "interrupt time" which is required for the hornet to fly and interrupt the hostile and the "danger situation" providing information whether the aircraft is a friend or hostile.

Each of the above-mentioned factors has different range of units such as the fuel (0 to 1000 litres), interrupt time (0 to 60 minutes), weapon status (0 to 100 %) and the danger situation (0 to 10 points). The following are two important decision selection rules, which were formulated using expert knowledge:

- The decision selection will have small value if the fuel used being too low, the interrupt time is too long, the hornet has low weapon status and the danger situation of FOE is high value.
- The decision selection will have high value if the fuel used being full, the interrupt time is fast enough, the hornet has high weapon status and the danger situation of FOE is low value.

In TACE, decision-making is always based on all states of all the decision factors. But sometime, a mission operator/commander can make a decision based on an important factor, such as the fuel used of the hornet is too low, enemy has more power weapon, quality and quantity of enemy aircraft. Table 2 shows the decision score at each stage of the TACE.

**Table 2. Some prior knowledge of the TACE**

| Fuel status (litres) | Interrupt time (minutes) | Weapon status (percent) | Danger situation (points) | Decision selection (points) |
|---|---|---|---|---|
| 0 | 60 | 0 | 10 | 0 |
| 100 | 55 | 15 | 8 | 1 |
| 200 | 50 | 25 | 7 | 2 |
| 300 | 40 | 30 | 5 | 3 |
| 400 | 35 | 40 | 4.5 | 4 |
| 500 | 30 | 60 | 4 | 5 |
| 600 | 25 | 70 | 3 | 6 |
| 700 | 15 | 85 | 2 | 7 |
| 800 | 10 | 90 | 1.5 | 8 |
| 900 | 5 | 96 | 1 | 9 |
| 1000 | 1 | 100 | 0 | 10 |

## 3. Soft Computing and Decision Trees

Soft computing paradigms are used to construct new generation intelligent hybrid systems consisting of neural networks, fuzzy inference system, approximate reasoning and derivative free optimisation techniques. It is well known that the intelligent systems, which can provide human like expertise such as domain knowledge, uncertain reasoning, and adaptation to a noisy and time varying environment, are important in tackling real world problems [26].

### 3.1 Artificial Neural Networks (ANNs)

Artificial neural networks have been developed as generalizations of mathematical models of biological nervous systems. Learning typically occurs by example through training, where the training algorithm iteratively adjusts the connection weights. In the Conjugate Gradient Algorithm (CGA) a search is performed along conjugate directions, which produces generally faster convergence than steepest descent directions. A search is made along the conjugate gradient direction to determine the step size, which will minimize the performance function along that line. A line search is performed to determine the optimal distance to move along the current search direction. Then the next search direction is determined so that it is conjugate to previous search direction. The general procedure for determining the new search direction is to combine the new steepest descent direction with the previous search direction. An important feature of the CGA is that the minimization performed in one step is not partially undone by the next, as it is the case with gradient descent methods. An important drawback of CGA is the requirement of a line search, which is computationally expensive. The Scaled Conjugate Gradient Algorithm (SCGA) [14] is basically designed to avoid the time-consuming line search at each iteration. SCGA combine the model-trust region approach, which is used in the Levenberg-Marquardt algorithm with the CGA.

### 3.2 Fuzzy Inference Systems (FIS)

Fuzzy inference system is a popular computing framework based on the concepts of fuzzy set theory, fuzzy if-then rules, and fuzzy reasoning. The basic structure of the fuzzy inference system consists of three conceptual components: a rule base, which contains a selection of fuzzy rules; a database, which defines the membership functions used in the fuzzy rule and a reasoning mechanism, which performs the inference procedure upon the rules and given facts to derive a reasonable output or conclusion. Most fuzzy systems employ the inference method proposed by Mamdani in which the rule consequence is defined by fuzzy sets and has the following structure [11]

$$If\ x\ is\ A_1\ and\ y\ is\ B_1\ then\ z_1 = C_1 \qquad (2)$$

Takagi, Sugeno and Kang proposed an inference scheme in which the conclusion of a fuzzy rule is constituted by a weighted linear combination of the crisp inputs rather than a fuzzy set and has the following structure [19]

$$If\ x\ is\ A_1\ and\ y\ is\ B_1, then\ z_1 = p_1 x + q_1 y + r \qquad (3)$$

Takagi-Sugeno FIS usually needs a smaller number of rules, because their output is already a linear function of the inputs rather than a constant fuzzy set [1].

### 3.3 Evolutionary Algorithms (EAs)

Evolutionary Algorithms are population based adaptive methods, which may be used to solve optimization problems, based on the genetic processes of biological organisms [6]. Over many generations, natural populations evolve according to the principles of natural selection and "Survival of the Fittest", first clearly stated by Charles Darwin in "On the Origin of Species". By mimicking this process, EAs are able to "evolve" solutions to real world problems, if they have been suitably encoded. The procedure may be written as the difference equation [6] as:

$$x[t+1] = s(v(x[t])) \qquad (4)$$

where $x(t)$ is the population at time $t$, $v$ is a random operator, and $s$ is the selection operator. The algorithm is as follows:

1. Generate the initial population $P(0)$ at random and set $i=0$;
2. Repeat until the number of iterations or time has reached or the population has converged.
   a. Evaluate the fitness of each individual in $P(i)$
   b. Select parents from $P(i)$ based on their fitness in $P(i)$
   c. Apply reproduction operators to the parents and produce offspring, the next generation, $P(i+1)$ is obtained from the offspring and possibly parents.

### 3.4 Neuro- Fuzzy Computing

Neuro Fuzzy (NF) computing is a popular framework for solving complex problems. If we have knowledge expressed in linguistic rules, we can build a FIS, and if we have data, or can learn from a simulation (training) then we can use ANNs. For building a FIS, we have to specify the fuzzy sets, fuzzy operators and the knowledge base. Similarly for constructing an ANN for an application the user needs to specify the architecture and learning algorithm. An analysis reveals that the drawbacks pertaining to these approaches seem complementary and therefore it is natural to consider building an integrated system combining the concepts. While the learning capability is an advantage from the viewpoint of FIS, the formation of linguistic rule base will be advantageous from the viewpoint of ANN [1]. In a fused NF architecture, ANN learning algorithms are used to determine the parameters of FIS. Fused NF systems share data structures and knowledge representations. A common way to apply a learning algorithm to a fuzzy system is to represent it in a special ANN like architecture. However the conventional ANN learning algorithms (gradient descent) cannot be applied directly to such a system as the functions used in the inference process are usually non differentiable. This problem can be tackled by using differentiable functions in the inference system or by not using the standard neural learning algorithm. Two neuro-fuzzy learning paradigms are presented in Section 4 and 5.

### 3.5 Classification and Regression Trees (CART)

Tree-based models are useful for both classification and regression problems. In these problems, there is a set of classification or predictor variables ($X_i$) and a dependent variable ($Y$). The $X_i$ variables may be a mixture of nominal and / or ordinal scales (or code intervals of equal-interval scale) and $Y$ a quantitative or a qualitative (i.e., nominal or categorical) variable [3] [18]. The CART methodology is technically known as binary recursive partitioning. The process is binary because parent nodes are always split into exactly two child nodes and recursive because the process can be repeated by treating each child node as a parent. The key elements of a CART analysis are a set of rules for splitting each node in a tree:

- deciding when a tree is complete; and
- assigning each terminal node to a class outcome (or predicted value for regression)

## 4 TACDSS Adaptation Using Takagi Sugeno FIS

We used the ANFIS framework [9] to develop the TACDSS based on a Takagi-Sugeno fuzzy inference system. The six-layered architecture of ANFIS is depicted in Figure 3.

Suppose there are two Input Linguistic Variables (ILV) $X$ and $Y$ and each ILV have three membership functions (MF) $A_1$, $A_2$ and $A_3$ and $B_1$, $B_2$ and $B_3$ respectively then a Takagi-Sugeno type fuzzy *if-then* rule could be set up as

Rule$_i$ : *If X is $A_i$ and Y is $B_i$ then $f_i = p_i X + q_i Y + r_i$*  (5)

Where $i$ is an index i = 1,2..n and $p$, $q$ and $r$ are the linear parameters.

Some layers of ANFIS have the same number of nodes and the nodes in the same layer have similar functions. Output of nodes in layer l is denoted as $O_{l,i}$ with l as the layer number and $i$ is neuron number of next layer. The function of each layer is described as follows.

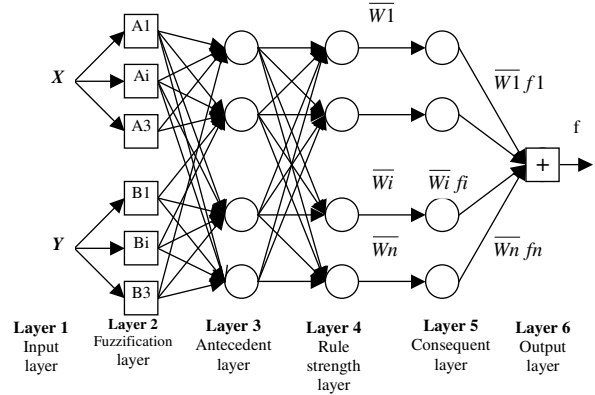

**Figure 3. Architecture of ANFIS**

**Layer 1**

The output of this node is the input values of the ANFIS

$O_{1,x} = x$
$O_{1,y} = y$  (6)

For TACDSS the four inputs are *fuel status*, *weapons inventory levels*, *time intercept* and the *danger situation*.

**Layer 2**

The output of nodes in this layer are presented as $O_{l,ip,i}$ where *ip* is the ILV and *m* is the degree of membership function of particular MF.

$O_{2,x,i} = \mu_{Ai(x)}$ or $O_{2,y,i} = \mu_{Bi(y)}$ for $i = 1,2$ and $3$ (7)

With three MFs for each input variable, "fuel status" has 3-membership functions: *full*, *half* and *low,* "time intercept" has *fast*, *normal* and *slow,* "weapon status" has *sufficient*, *enough* and *insufficient* and the "danger situation" has *very dangerous*, *dangerous* and *endanger*.

**Layer 3**

The output of nodes in this layer is the product of all the incoming signals which denotes

$O_{3,n} = W_n = \mu_{Ai}(x) \times \mu_{Bi}(y)$ (8)

where $i = 1,2$ and $3$, $n$ is the number of fuzzy rule. In general, any T-norm operator will perform the fuzzy 'AND' operation in this layer. With 4 ILV and 3 MFs for each input variable the TACDSS will have 81 ($3^4 = 81$) fuzzy *if-then* rules.

**Layer 4**

The node in this layer calculates the ratio of the $i^{th}$ fuzzy Rules Firing Strength (RFS) to the sum of RFS.

$O_{4,n} = \overline{w_n} = \dfrac{w_n}{\sum\limits_{n=1}^{81} w_n}$ where $n = 1,2,...,81$ (9)

The number of nodes in this layer is the same as the number of nodes in layer 3. The output of this layer is also called *normalized firing strength*.

**Layer 5**

The node in this layer is an adaptive node defined as:

$O_{5,n} = \overline{w_n} f_n = \overline{w_n}(p_n x + q_n y + r_n)$ (10)

$p_n$, $q_n$, $r_n$ are the rule *consequent parameters*. This layer also has the same number of nodes in layer 4 (81 numbers).

**Layer 6**

The single node in this layer is responsible for the defuzzification process using the center of gravity technique to compute the overall output as the summation of all the incoming signals.

$O_{6,1} = \sum\limits_{n=1}^{81} \overline{w_n} f_n = \dfrac{\sum\limits_{n=1}^{81} w_n f_n}{\sum\limits_{n=1}^{81} w_n}$ (11)

ANFIS makes use of a mixture of backpropagation to learn the premise parameters and least mean square estimation to determine the consequent parameters. A step in the learning procedure has two parts: In the first part the input patterns are propagated, and the optimal conclusion parameters are estimated by an iterative least mean square procedure, while the antecedent parameters (membership functions) are assumed to be fixed for the current cycle through the training set. In the second part the patterns are propagated again, and in this epoch, back propagation is used to modify the antecedent parameters, while the conclusion parameters remain fixed. This procedure is then iterated [9]. Some details are given below.

ANFIS output $f = O_{6,1} = \dfrac{w_1}{\sum\limits_n wn} f_1 + \dfrac{w_2}{\sum\limits_n wn} f_2 + ... + \dfrac{wn}{\sum\limits_n wn} f_n$ (12)

$= \overline{w_1}(p_1 x + q_1 y + r_1) + \overline{w_2}(p_2 x + q_2 y + r_2) + ... + \overline{w_n}(p_n x + q_n y + r_n)$ (13)

$= (\overline{w_1} x)p_1 + (\overline{w_1} y)q_1 + \overline{w_1} r_1 + (\overline{w_2} x)p_2 + (\overline{w_2} y)q_2 + \overline{w_2} r_2 + ... + (\overline{w_n} x)p_n + (\overline{w_n} y)q_n + \overline{w_n} r_n$ (14)

where $n$ is the number of nodes in layer 5. From this, the output can be rewritten as

$f = F(i,S)$ (15)

where $F$ is a function, $i$ is the vector of input variables and $S$ is a set of total parameters of consequent of $n^{th}$ fuzzy rule. If there exists a composite function $H$ such that $H \circ F$ is a linear in some elements of $S$, then these elements can be identified by the least square method. If the parameter set is divided into two sets of $S_1$ and $S_2$ which is defined as

$S = S_1 \oplus S_2$ (16)

Where $\oplus$ represents direct sum, such that $H \circ F$ is linear in the elements of $S_2$, the function $f$ can be known as

$H(f) = H \circ F(I,S)$ (17)

Givens values of $S_1$, substitute the $S$ training data into the equation $H(f)$. $H(f)$ can be written as Matrix equation of $AX = Y$.

where $X$ is an unknown vector whose elements are parameters in $S_2$.

If $|S_2| = M$ ($M$ is number of linear parameters) then the dimension of matrix $A$, $X$ and $Y$ are $P \times M$, $M \times 1$ and $P \times 1$ respectively. This is a standard linear least-squares problem and the best solution of $X$ that minimizes $||AX - Y||^2$ is the least square estimate (LSE) $X^*$

$X^* = (A^T A)^{-1} A^T Y$ (18)

Where $A^T$ is the transpose of $A$, $(A^T A)^{-1} A^T$ is the pseudo inverse of $A$ if $A^T A$ is a non-singular. Let the $i^{th}$ row vector of matrix $A$ be $a_i^T$ and the $i^{th}$ element of $Y$ be $y_i^T$, then $X$ can be calculated as

$X_{i+1} = X_i + S_{i+1} a_{i+1}(y_i^T - y_{i+1}^T - a_{i+1}^T X_i)$ (19)

$S_{i+1} = S_i - \dfrac{S_i a_{i+1} y_{i+1}^T - S_i}{1 + a_{i+1}^T S_i a_{i+1}}$, $I = 0,1,..., P - 1$ (20)

The LSE $X^*$ is equal to $X_p$. The initial condition of $X_{i+1}$ and $S_{i+1}$ are $X_0 = 0$ and $S_0 = \gamma I$ where $\gamma$ is positive large number and $I$ is identity matrix of dimension $M \times M$.

When hybrid learning is applied in the batch mode, each epoch is composed of a forward pass and a backward pass. In the forward pass, the node output $I$ of each layer is calculated until the corresponding matrices $A$ and $Y$ are obtained. The parameters of $S_2$ are identified by the pseudo inverse equation as mentioned above. After the parameters of $S_2$ are obtained, the process will compute the error measure for each training data pair. In the backward pass, the error signals, that is the derivative of the error measure with respect to each node output, propagates from the output to the input end. At the end of the backward pass, the parameter $S_1$ is updated by the steepest descent method as follows:

$$\alpha = -\eta \frac{\partial E}{\partial \alpha} \qquad (21)$$

where $\alpha$ is a generic parameter and $\eta$ is a learning rate and $E$ is an error measure.

$$\eta = \frac{k}{\sqrt{\Sigma \alpha (\frac{\partial E}{\partial \alpha})^2}} \qquad (22)$$

where $k$ is the step size.

For given fixed values of parameters in $S_1$, the parameters in $S_2$ are guaranteed to be global optimum point in the $S_2$ parameters space due to the choice of the squared error measure. This hybrid learning method can decrease the dimension of the search space using the steepest descent method. This method can reduce the time need to reach convergence. The step size $k$ will influence the speed of convergence. The observation shows that if $k$ is small, the gradient method will closely approximate the gradient path. The convergence will be slow since the gradient being calculated many times. If the step size $k$ is large, convergence will initially very fast. Based on these observations the step size $k$ is updated by the two heuristic rules.

- If $E$ undergoes 4 continuous reductions then increase $k$ by 10%
- If $E$ undergoes continuous combination of increase and decrease then reduce $k$ by 10%

## 5 TACDSS Adaptation Using Mamdani FIS

We have made use of the Fuzzy Neural Network (FuNN) framework [10] for learning the Mamdani fuzzy inference method. A functional block diagram of the FuNN model is depicted in Figure 4 and it consists of two phases of learning.

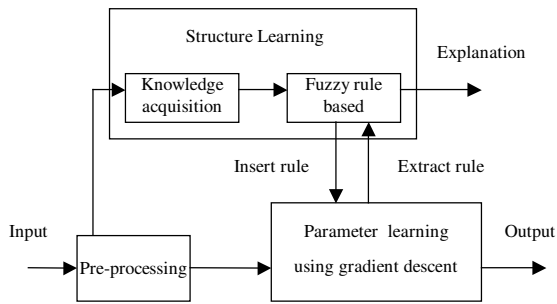

**Figure 4. A general schematic diagram of FuNN**

The first phase is the structure learning (*if-then* rules) using the knowledge acquisition module. The second phase is the parameter learning for tuning membership functions to achieve a desired level of performance. FuNN uses a gradient descent-learning algorithm to fine-tune the parameters of the fuzzy membership functions. In the connectionist structure, the input and output nodes represent the input states and output control-decision signals, respectively, and in the hidden layers, there are nodes functioning as quantification of membership functions (MFs) and *if-then* rules. We used a simple and straightforward method proposed by Wang and Mendel [24] for generating fuzzy rules from numerical input-output training data. The task here is to generate a set of fuzzy rules from the desired input-output pairs and then use these fuzzy rules to determine the complete structure of the TACDSS.

Suppose we are given the following set of desired input - ($x_1$, $x_2$) and output ($y$) data pairs ($x_1$, $x_2$, $y$): (0.6, 0.2; 0.2), (0.4, 0.3; 0.4). In TACDSS, input variable *fuel used* has a degree of 0.8 in *half*, a degree of 0.2 in *full*. Similarly, input variable *time intercept* has degree of 0.6 in *empty* and of 0.3 in *normal*. Secondly, assign $x_1^i$, $x_2^i$, and $y^i$ to a region that has maximum degree. Finally, obtain one rule from one pair of desired input-output data, for example,
($x_1^1$, $x_2^1$, $y^1$) => [$x_1^1$ (0.8 in *half*), $x_2^1$ (0.2 in *fast*), $y^1$ (0.6 in *acceptable*)],

$R_1$: if $x_1$ is *half* and $x_2$ is *fast*, then $y$ is *acceptable*

($x_1^2$, $x_2^2$, $y^2$), => [$x_1$(0.8 in *half*), $x_2$(0.6 in *normal*), $y^2$(0.8 in *acceptable*)],

$R_2$: if $x_1$ is *half* and $x_2$ is *normal*, then $y$ is *acceptable*

Assign a degree to each rule. To resolve a possible conflict problem, that is, rules having the same antecedent but a different consequent, and to reduce the number of rules, we assign a degree to each rule generated from data pairs and accept only the rule from a conflict group that has a maximum degree. In other words, this step is performed to delete redundant rules, and therefore obtain a concise fuzzy rule base. The following product strategy is used to assign a degree to each rule. The degree of the rule denoted by

$Ri$ : if $x_1$ is $A$ and $x_2$ is $B$, then $y$ is $C(w_i)$

The rule weight is defined as

$w_i = \mu_A(x_1)\mu_B(x_2)\mu_c(y)$

For example in the TACE, $R_1$ has a degree of

$W_1 = \mu_{half}(x_1)\,\mu_{fast}(x_2)\,\mu_{acceptable}(y) = 0.8 \times 0.2 \times 0.6 = 0.096$

and $R_2$ has a degree of

$W2 = \mu_{half}(x_1)\,\mu_{normal}(x_2)\,\mu_{acceptable}(y) = 0.8 \times 0.6 \times 0.8 = 0.384$

Note, that if two or more generated fuzzy rules have the same preconditions and consequents, then the rule that has maximum degree is used. In this way, assigning the degree to each rule, the fuzzy rule base can be adapted or updated by the relative weighting strategy: the more task related the rule becomes, the more weight degree the rule gains. As a result, not only is the conflict problem resolved, but also the number of rules is reduced significantly. After the structure-learning phase (*if-then* rules), the whole network structure is established, and the network enters the second learning phase to optimally adjust the parameters of the membership functions using a gradient descent learning algorithm to minimise the error function

$$E = \frac{1}{2} \sum_{x} \sum_{l=1}^{q} (d_1 - y_l)^2 \quad (23)$$

where *d* and *y*, are the target and actual outputs for an input *x*. This approach is very much similar to the *MF* parameter tuning in ANFIS.

### 5.1 Membership Function Parameter Optimisation Using EAs

We have investigated the usage of evolutionary algorithms (EAs) to optimise the number of rules and fine-tune the membership functions [8][12][21]. A simple way to optimise membership function parameters is to represent only the parameter showing the centre of MF's to speed up the adaptation process and to reduce spurious local minima over the center and width.

The GA module for adapting FuNN is designed as a stand-alone system for optimising the MF's if the rules are already available. Both antecedent and consequent MF's are optimised. Chromosomes are represented as strings of floating-point numbers, rather than strings of bits. In addition, mutation of a gene is implemented as a re-reinitialisation of the gene, rather than an alteration of the existing allele. Figure 5 shows the chromosome structure showing the input and output MF parameter structure. One point crossover is used for the reproduction of chromosomes.

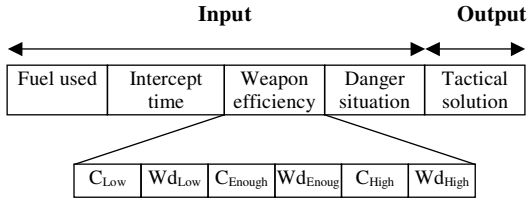

**Figure 5. Chromosome representation**

### 6. Experimentation Results for Developing the TACDSS

Our master data set comprises of 1000 numbers. To avoid any bias on the data, from the master dataset, we randomly created two sets of training (Dataset A - 90% and Dataset B- 80%) and test data (10% and 20%). All the experimentations were repeated three times and the average errors are reported.

### 6.1 Takagi Sugeno FIS

In addition to the development of the Takagi Sugeno FIS, we also investigated the behaviour of TACDSS for different membership functions (shape and quantity per ILV). We also explored the importance of different learning methods for fine-tuning the rule antecedents and consequents. Keeping the consequent parameters constant, we fine-tuned the membership functions alone using the gradient descent technique (backpropagation). Further, we used the hybrid learning method wherein the consequent parameters were also adjusted according to the least squares algorithm. Even though back-propagation is faster than the hybrid technique, learning error and the decision scores were better for the latter technique. We used 3 Gaussian MFs for each ILV. Figure 6 (a) and (b) shows the three MFs for the ILV "fuel used" before and after training. The consequent parameters of fuzzy rule before training was set to zero and the parameters were learned using the hybrid learning approach.

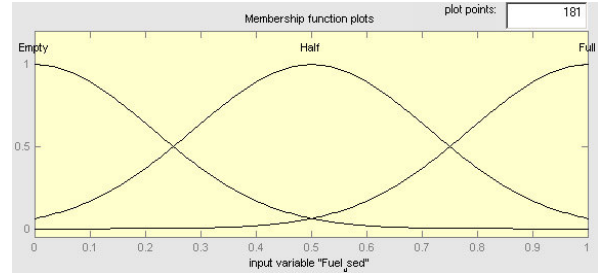

**(a)**

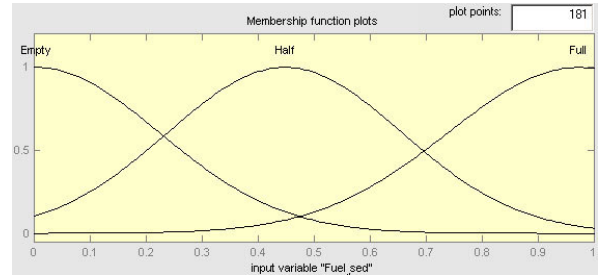

**(b)**

**Figure 6. The MFs of the ILF "fuel used" (a) before and (b) after learning**

**Comparison of the shape of membership functions**

In this section, we will demonstrate the importance of the shape of membership functions. We used the hybrid-learning technique and each ILV has three MFs. Table 3 shows the convergence of the training RMSE during the 15 epochs learning using four different membership functions for 90% and 80% training data. 81 fuzzy if-then rules were created initially using a grid-partitioning algorithm. We considered Generalised bell, Gaussian, trapezoidal and isosceles triangular membership functions. Figure 7 illustrates the training convergence curve for different MF's.

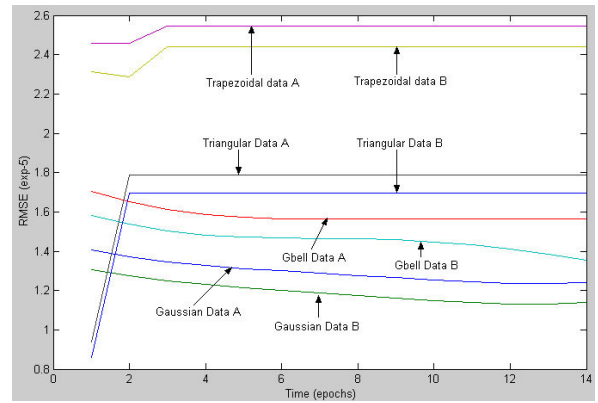

**Figure 7. Training error for the different MF's**

As evident from Table 3 and Figure 7, the lowest training and testing error was obtained using Gaussian MF.

**Table 3. Learning performance for different MF's**

| Training Root Mean Squared Error (E- 05) | | | | | | | | |
|---|---|---|---|---|---|---|---|---|
| Epochs | Gaussian | | Gbell | | Trapezoidal | | Triangular | |
| | Data A | Data B | Data A | Data B | Data A | Data B | Data A | Data B |
| 1 | 1.406 | 1.305 | 1.706 | 1.581 | 2.459 | 2.314 | 0.9370 | 0.8610 |
| 2 | 1.372 | 1.274 | 1.652 | 1.537 | 2.457 | 2.285 | 1.789 | 1.695 |
| 3 | 1.347 | 1.249 | 1.612 | 1.505 | 2.546 | 2.441 | 1.789 | 1.695 |
| 4 | 1.328 | 1.230 | 1.586 | 1.483 | 2.546 | 2.441 | 1.789 | 1.695 |
| 5 | 1.312 | 1.214 | 1.571 | 1.471 | 2.546 | 2.441 | 1.789 | 1.695 |
| 6 | 1.300 | 1.199 | 1.565 | 1.466 | 2.546 | 2.441 | 1.789 | 1.695 |
| 7 | 1.288 | 1.186 | 1.564 | 1.465 | 2.546 | 2.441 | 1.789 | 1.695 |
| 8 | 1.277 | 1.173 | 1.565 | 1.464 | 2.546 | 2.441 | 1.789 | 1.695 |
| 9 | 1.265 | 1.160 | 1.565 | 1.459 | 2.546 | 2.441 | 1.789 | 1.695 |
| 10 | 1.254 | 1.148 | 1.565 | 1.448 | 2.546 | 2.441 | 1.789 | 1.695 |
| 11 | 1.243 | 1.138 | 1.565 | 1.431 | 2.546 | 2.441 | 1.789 | 1.695 |
| 12 | 1.236 | 1.132 | 1.565 | 1.409 | 2.546 | 2.441 | 1.789 | 1.695 |
| 13 | 1.234 | 1.132 | 1.565 | 1.384 | 2.546 | 2.441 | 1.789 | 1.695 |
| 14 | 1.238 | 1.138 | 1.565 | 1.355 | 2.546 | 2.441 | 1.789 | 1.695 |
| Test Root Mean Squared Error (E- 05) | | | | | | | | |
| | 1.44 | 1.22 | 1.78 | 1.36 | 2.661 | 2.910 | 1.858 | 1.858 |

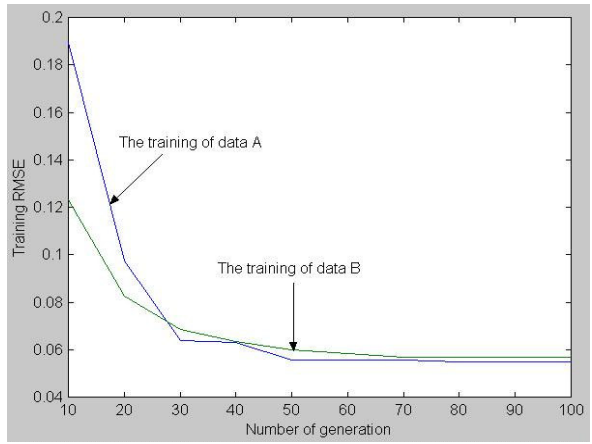

**Figure 8. Convergence of training using EA's**

### 6.2 Mamdani Fuzzy Inference System

We used the FuzzyCOPE [25] to investigate the tuning of membership functions using backpropagation and evolutionary algorithms. The learning rate and momentum were set at 0.5 and 0.3 respectively for 10 epochs. We obtained training RMSE 0.2865 (Data A) and 0.2894 (Data B). We further improved the training performance using genetic algorithms. We used a population size = 50, number of generations = 100 and mutation rate = 0.01

We used the tournament selection strategy and Figure 8 illustrates the learning convergence during the 100 generations for Datasets A and B. 54 fuzzy *if-then* rules were extracted after the learning process. Table 4 summarizes the training and test performances.

**Table 4. Performance of Mamdani FIS using EAs**

| Root Mean Squared Error (RMSE) | | | |
|---|---|---|---|
| Data A | | Data B | |
| Training | Test | Training | Test |
| 0.0548 | 0.0746 | 0.0567 | 0.0612 |

### 6.3 Neural Networks

We used 30 hidden neurons for Data A and 32 hidden neurons for Data B. We used a trial and error approach to finalize the architecture of the neural network. We used the scaled conjugate gradient algorithm to develop the TACDSS. The training was terminated after 1000 epochs. Figure 9 depicts the convergence of training during 1000 epochs learning. Table 5 summarizes the training and test performances.

### 6.4 Classification and Adaptive Regression Trees

We used the CART [5] simulation environment to develop the decision trees. We selected the minimum cost tree regardless the size of the tree. Figures 10 and 11 illustrate the variation of error with reference to the number of terminal nodes for Datasets A and B. For Data A, the developed tree has 122 terminal nodes as shown in Figure 12 while for Data B the resting tree had 128 terminal nodes as depicted in Figure 13. Training and test performances are summarized in Table 5.

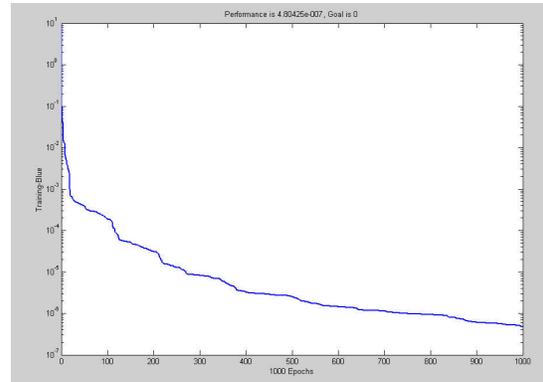

**Figure 8. Neural network training using SCGA**

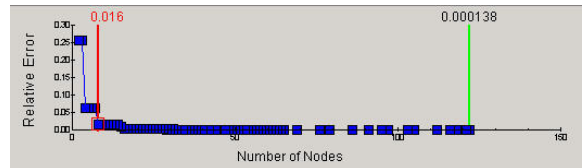

**Figure 10. Dataset A: Variation of relative error**

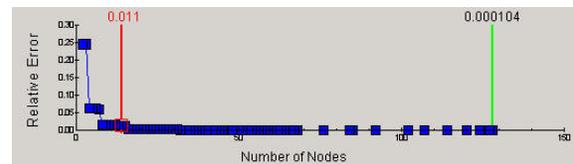

**Figure 11. Dataset B: Variation of relative error**

Figure 14 illustrates the performance comparison of the different intelligent paradigms for developing the TACDSS. For clarity we have chosen only 20% of the test results of Dataset B.

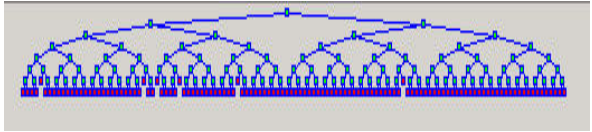

**Figure 12. Dataset A: decision tree with 122 nodes**

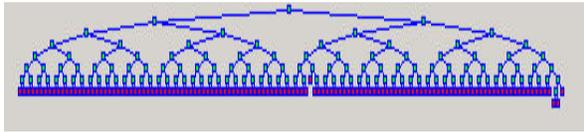

**Figure 13. Dataset B: decision tree with 128 nodes**

**Table 5. Performance of ANN and decision trees**

|  | Data A | | Data B | |
|---|---|---|---|---|
|  | Training | Testing | Training | Testing |
|  | RMSE | | | |
| CART | 0.00239 | 0.00319 | 0.00227 | 0.00314 |
| ANN | 0.00105 | 0.00095 | 0.00041 | 0.00062 |

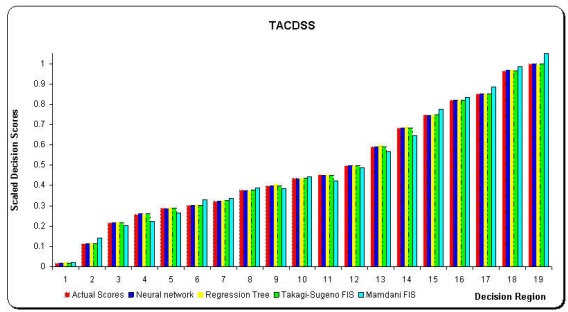

**Figure 14. Performance of the different intelligent paradigms.**

## 7. Conclusions and Discussions

In this paper, we have presented different soft computing and machine learning paradigms for developing a tactical air combat decision support system. The techniques explored were a Takagi Sugeno fuzzy inference system learned using neural network learning techniques, a Mamdani fuzzy inference system learned using evolutionary algorithms and neural network learning, feed-forward neural network trained using the scaled conjugate gradient algorithm and classification and adaptive regression trees.

Our experimentation results using two different datasets reveal the importance of fuzzy inference systems to construct accurate decision support systems. As expected, by providing more training data (90% of the randomly choosed master data set), the models were able to learn and generalize more accurately. Takagi-Sugeno fuzzy inference system has the lowest RMSE on both test datasets. Since learning involves a complicated procedure the training process of the Takagi-Sugeno fuzzy inference system took longer time while compared to Mamdani fuzzy inference method. Hence there is a compromise between performance and the computational complexity (training time). Our experiments using different membership function shapes also reveal that Gaussian membership function is the best shape for the construction of accurate decision support systems. The proposed neural network could outperform Mamdani fuzzy inference system and CART. Two important features of the developed classification and regression tree are its easy interpretability and least complexity. Due to the one pass training approach; the CART algorithm also has the lowest computational load.

The empirical results clearly demonstrate that the proposed techniques are reliable and could be used for constructing more complicated decision support systems. Experiments on the two independent data sets also reveal that the techniques are not biased on the data itself. It may also be concluded that fusing different intelligent systems knowing their strengths and weakness could help to mitigate the limitations and take advantage of the opportunities to produce more efficient decision support systems than those could be built with stand alone systems.

Our future work will be directed towards optimisation of the different intelligent paradigms [2], which we have already used and also to develop new adaptive reinforcement learning systems that can update the knowledge from data especially when no expert knowledge is available.